\documentclass[journal]{IEEEtran}
\usepackage{amsfonts,dsfont}
\usepackage{balance}
\usepackage{booktabs}
\usepackage{amsmath}
\usepackage{tabularx}
\usepackage{subfigure}
\usepackage{graphicx}
\usepackage{multirow}
\usepackage{rotating}
\usepackage{algorithm}
\usepackage{algorithmic}
\usepackage{url}
\usepackage{amssymb}
\usepackage{amsthm}
\usepackage{enumerate}
\usepackage{color}
\usepackage{bm}

\renewcommand{\algorithmicrequire}{\textbf{Input:}}
\renewcommand{\algorithmicensure}{\textbf{Output:}}

\newcommand{\Rmnum}[1]{\expandafter\@slowromancap\romannumeral #1@}

\begin{document}
	\makeatletter
	\makeatother
	\title{Dual Information Enhanced Multi-view Attributed Graph Clustering}
	
	\author{Jia-Qi~Lin,
		Man-Sheng~Chen, Xi-Ran~Zhu,
		Chang-Dong~Wang,~\IEEEmembership{Member,~IEEE,}
		Haizhang~Zhang
		\thanks{Jia-Qi Lin and Haizhang Zhang are with School of Mathematics (Zhuhai), Sun Yat-sen University, Zhuhai, China. E-mail: linjq56@mail2.sysu.edu.cn, zhhaizh2@mail.sysu.edu.cn.}
		\thanks{Man-Sheng Chen, Xi-Ran~Zhu and Chang-Dong Wang are with School of Computer Science and Engineering, Sun Yat-sen
			University, Guangzhou, China, Key Laboratory of Machine Intelligence and Advanced Computing, Ministry of Education, China, and Guangdong Province Key Laboratory of Computational Science, Guangzhou, China.
			E-mail: chenmsh27@mail2.sysu.edu.cn, zhuxr3@mail2.sysu.edu.cn,
			changdongwang@hotmail.com.}
		\thanks{Corresponding author: Chang-Dong Wang}
	}
	
	\markboth{Journal of \LaTeX\ Class Files,~Vol.~14, No.~8, August~2015}%
	{Shell \MakeLowercase{\textit{et al.}}: Bare Demo of IEEEtran.cls for IEEE Journals}

	\maketitle

	\begin{abstract}
		Multi-view attributed graph clustering is an important approach to partition multi-view data based on the attribute feature and adjacent matrices from different views.
		Some attempts have been made in utilizing Graph Neural Network (GNN), which have achieved promising clustering performance. Despite this, few of them pay attention to the inherent specific information embedded in multiple views. Meanwhile, they are incapable of recovering the latent high-level representation from the low-level ones, greatly limiting the downstream clustering performance. To fill these gaps, a novel Dual Information enhanced multi-view Attributed Graph Clustering~(DIAGC) method is proposed in this paper.
		Specifically, the proposed method introduces the Specific Information Reconstruction~(SIR) module to disentangle the explorations of the consensus and specific information from multiple views, which enables GCN to capture the more essential low-level representations. Besides, the Mutual Information Maximization~(MIM) module maximizes the agreement between the latent high-level representation and low-level ones, and enables the high-level representation to satisfy the desired clustering structure with the help of the Self-supervised Clustering~(SC) module.
		Extensive experiments on several real-world benchmarks demonstrate the effectiveness of the proposed DIAGC method compared with the state-of-the-art baselines.
	\end{abstract}
	
	
	\section{Introduction}
	\IEEEPARstart{C}{lustering} is one of the most fundamental tasks of unsupervised learning~\cite{doi:10.1126/science.1136800,DBLP:journals/tkde/SongNW13,DBLP:journals/tip/ChatterjeeM09,DBLP:conf/kdd/FisherCWR15,DBLP:journals/jsac/DaiW17,DBLP:conf/sigir/ZhangCJWH0Y22,DBLP:journals/dase/ChenLLLWHL22}. The basic idea is to divide the data into several disjoint clusters, where the samples in the same cluster are highly similar to each other, and the samples in different clusters are less related. With the rapid growth of social media and the internet, data has become more complex since it is collected, processed, and stored in various ways. For example, on the movie network data, a movie can be connected with other movies not only with the co-actor relationship but also with the co-director relationship. Besides, each movie has its own keywords as the attribute features. This kind of data is termed as multi-view attributed graph data~\cite{DBLP:conf/ijcai/LinK21}, which contains one attribute feature matrix and multiple adjacent matrices.
	
	\begin{figure}[!t]
		\vskip -0.1in
		\renewcommand{\subfigcapskip}{0pt}
		\renewcommand{\subfigbottomskip}{0pt}
		\centerline
		{
			{\subfigure
				{\includegraphics[width=0.25\textwidth]{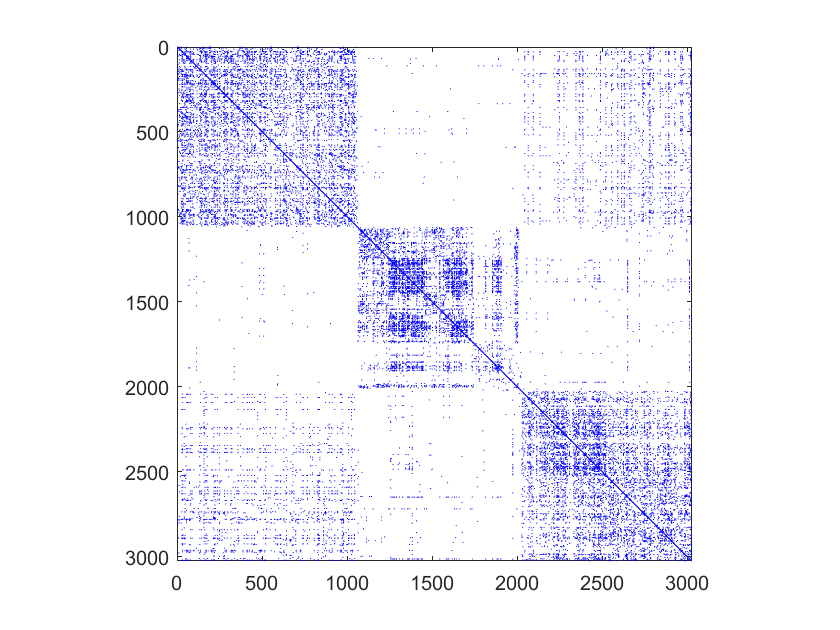}
					\label{fig:acm_1}
				}	
			}
			\hspace{-10mm}
			{\subfigure
				{\includegraphics[width=0.25\textwidth]{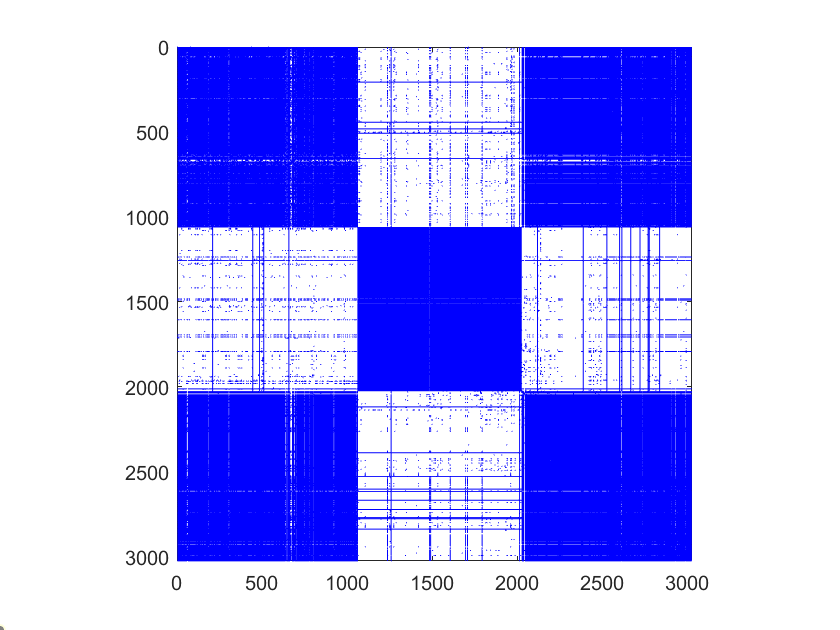}
					\label{fig:acm_2}
				}
			}
			
		}
		\caption{The adjacent matrices of the ACM dataset. It can be seen that the graph structures of different views are distinct, which indicates that the specific information is common in real-world data and should not be neglected in the multi-view attributed graph clustering method.}
		\label{fig:acm_ajdacent}
	\end{figure}
	
	Since the adjacent matrix reflects the topological structure of graph data, numerous graph-based multi-view clustering methods~\cite{DBLP:conf/acml/ShuL15,DBLP:conf/ijcai/NieLL16,DBLP:journals/tkde/HouNTY17} have been developed to explore such information, e.g., multi-view spectral clustering~\cite{DBLP:conf/nips/KumarRD11,DBLP:conf/aaai/XiaPDY14,DBLP:journals/tkde/WangLY16,DBLP:journals/tnn/ChenXHZ22}, graph-based multi-view subspace clustering~\cite{DBLP:journals/tkde/ZhangLXZZZC22,DBLP:journals/tip/WangLZZZGZ22,DBLP:conf/aaai/LiuW00LZG22,DBLP:journals/isci/LvKWJX21,DBLP:journals/tmm/XiaoGHC21,DBLP:journals/tnn/WangCHLY21}, tensor learning-based multi-view clustering~\cite{DBLP:conf/aaai/GaoXWXZ20,DBLP:journals/tip/ChenWPHZ21} and \textit{ect}.. The multi-view spectral clustering methods~\cite{DBLP:journals/pami/LiuZLWTYSWG19,DBLP:conf/icml/0001HLZZ19,DBLP:journals/tnn/ZhouDLSFL21} are derived from the single-view spectral clustering methods, which maximize the agreements between different views and
	explicitly employ the normalized cut on the graph for a soft assignment fitting views.
	The graph-based multi-view subspace clustering methods~\cite{DBLP:conf/ijcai/NieLL17,DBLP:journals/tkde/WangYL20,DBLP:journals/tcyb/LiuHWFY21,9843949,9597486} are based on the assumption that all views are shared with one latent embedding. They construct the self-representation matrices for all views and perform the single-view clustering method, e.g., $k$-means, on the consensus representation matrix for the cluster partition.
	Moreover, tensor learning-based multi-view clustering methods~\cite{DBLP:journals/tmm/ChenWXLHZ22,DBLP:journals/tcsv/ChenXPLZ22,DBLP:journals/isci/FuYCZ22,9926008} also show promising clustering performance on multi-view graph data, which stack the representation matrices into a three-order tensor to explore the high-order correlation among all views. Despite great success, they only consider the topological structures, i.e, the adjacent matrices of the graph data, and pay little attention to the node attribute information~\cite{DBLP:journals/nn/XiaWYGHG22}.
	
	To fully explore the topological information and the node attribute information of graph data, some efforts have been made in the Graph Neural Network~(GNN)-based clustering methods in recent years~\cite{DBLP:conf/nips/SureshLHN21}. Compared with the aforementioned shallow models, GNN has powerful nonlinear feature extraction ability, and the features extracted by GNN can directly be applied to the downstream tasks, e.g. clustering.
	The Graph Convolutional Network~(GCN)-based methods~\cite{DBLP:conf/www/Bo0SZL020} define a convolutional operation on the graph data for feature extraction.
	Motivated by the great success of autoencoders, Graph Auto-Encoder~(GAE) network-based methods~\cite{DBLP:journals/corr/KipfW16a} extract deep embeddings in a self-training manner.
	The GNN-based multi-view clustering methods consider the complementary information between different views by extracting the deep representation of each view, and fusing them into the consensus one to mine the essential information for clustering.
	
	Despite significant performance, there are two main issues to be addressed for the existing deep multi-view attributed graph clustering methods:
	(1) Existing methods do not consider the inherent specific information embedded in each view when computing the consensus representation, which may interfere with the performance of the downstream tasks, i.e., clustering. The purpose of the GNN-based multi-view attributed graph clustering method is to find a cluster partition from the consensus representation that fits all views. For multi-view attributed graph data, as illustrated in \figurename~\ref{fig:acm_ajdacent}, the adjacency matrix of each view has its own special structure, but this structure information could not be dealt with appropriately, resulting in degraded clustering performance.
	(2)~They use low-level representations to construct the consensus representation, which may introduce meaningless information and affect the downstream clustering task. The existing methods obtain the cluster partition directly from the low-level representations of each view, and some of them reconstruct the multiple adjacent matrices from one consensus representation. The target of multi-view clustering is to find a consensus cluster partition from all views while the reconstruction process aims to reconstruct the diversity of different views from the same representation, which induces the conflict between the two tasks and leads to sub-optimal clustering results.
	
	To solve the aforementioned problems, in this paper, a novel Dual Information enhanced multi-view Attributed Graph Clustering~(DIAGC) method is proposed, where the explorations of the consensus and specific information are disentangled elegantly. Besides, it fully utilizes the non-linearity of the low-level representations to generate the latent high-level clustering-oriented representation.
	Specifically, the proposed DIAGC method employs the Mutual Information Maximization~(MIM) module and Specific Information Reconstruction~(SIR) module to mine the rich information embedded in the deep representations of each view for the latent high-level representation. Besides, to avoid interferences from the inherent specific information embedded in different views, the Specific Information Reconstruction~(SIR) module is introduced in DIAGC, enabling the GCN encoder to extract the purer consensus representation. The Self-supervised Clustering~(SC) module forces the latent high-level representation to be a clustering-oriented one.
	By training these modules simultaneously, the modules can boost each other for better clustering results in a mutual manner.
	
	In general, the contributions of this paper can be summarized as follows:
	\begin{enumerate}
		\item To the best of our knowledge, this is the first attempt to disentangle the consensus and specific information learning for deep multi-view attributed graph clustering.
		\item A novel Dual Information enhanced multi-view Attributed Graph Clustering~(DIAGC) method is developed, where the latent high-level clustering-oriented representation learning as well as the specific information reconstruction are seamlessly integrated into a unified framework.
		\item The experimental results on several real-world benchmarks compared with the state-of-the-art baselines demonstrate the effectiveness of the proposed DIAGC method.
	\end{enumerate}
	
	The organization of this paper is as follows. In Section~\ref{sec:RelatedWork}, the related works are outlined. In Section~\ref{sec:basic notation}, some basic notations and the problem definitions are briefly reviewed. Section~\ref{sec:The Proposed Method} introduces the proposed DIAGC method in detail. The experimental setups, real-world benchmarks, state-of-the-art baselines and experimental results are presented and analyzed in Section~\ref{sec:Experiments}. The conclusion of this paper is given in Section~\ref{sec:Conclusion}.	
	
	\section{Related Work}
	\label{sec:RelatedWork}
	\subsection{Multi-view Clustering}
	The basic idea of multi-view clustering is to explore the consensus and complementary information embedded in multi-view data and learn a cluster partition that fits all views.
	Multi-view spectral clustering methods aim to learn a soft cluster indicator matrix directly from the original data.
	For instance, Kumar~\textit{et al.}~\cite{DBLP:conf/nips/KumarRD11} employ the co-regularized scheme to maximize the agreement between views. Xia~\textit{et al.}~\cite{DBLP:conf/aaai/XiaPDY14} propose a robust Markov-chain-based multi-view spectral clustering method where a low-rank transition matrix calculated by graphs is regarded as the input.
	Wang~\textit{et al.}~\cite{DBLP:journals/tkde/WangLY16} consider the clustering quality of each view and the clustering results agreement between different views.
	In~\cite{DBLP:conf/ijcai/NieLL17}, the self-weighting learning strategy is introduced to learn a block diagonal matrix for clustering from multiple graph inputs.
	In~\cite{DBLP:journals/tkde/WangYL20}, similarity matrix learning, unified graph learning and indicator learning are incorporated into a unified framework. Liu~\textit{et al.}~\cite{DBLP:journals/tcyb/LiuHWFY21} consider the intra-view and inter-view correlations together and propose an adaptively weighted multi-view spectral clustering method.
	For graph-based multi-view subspace clustering methods, the self-representation matrices are constructed from the original graph data for clustering. In~\cite{DBLP:conf/cvpr/CaoZFLZ15}, the Hilbert Schmidt Independence Criterion~(HSIC) term is introduced to explore the complementary information between the self-representation matrices from different views. Tang~\textit{et al.}~\cite{DBLP:journals/tmm/TangZLLWZW19} construct a consensus self-representation matrix with low-rank constraints for clustering. Zhang~\textit{et al.}~\cite{DBLP:journals/kbs/ZhangZHWH20} propose a one-step framework to explore the intra-view structure and the inter-view consistency simultaneously.
	In~\cite{DBLP:journals/tcyb/ZhouZPBY20}, the underlying correlations among different views and the specific information are explored to strengthen clustering performance.
	Recently, tensor learning-based multi-view clustering methods have achieved great progress. Xie~\textit{et al.}~\cite{DBLP:journals/ijcv/XieTZLZQ18} stack the self-representation matrices from different views into a three-order tensor, and impose the t-SVD-based Tensor Nuclear Norm~(TNN) on it to explore the high-order correlation among all views. Gao~\textit{et al.}~\cite{DBLP:conf/aaai/GaoXWXZ20} utilize the prior knowledge of the singular values and propose a weighted TNN for multi-view clustering. Wu~\textit{et al.} extend~\cite{DBLP:conf/aaai/XiaPDY14} by stacking the transition matrices into a tensor with the TNN constraint~\cite{DBLP:journals/tip/WuLZ19}. In~\cite{9714775}, Chen~\textit{et al.} employ the weighted TNN for multi-view spectral clustering.
	
	Unfortunately, the above methods can only process the attribute feature matrices or adjacent matrices separately, which can not fully explore the properties of the multi-view attributed graph data. To fill this gap, Lin~\textit{et al.}~\cite{DBLP:conf/ijcai/LinK21} introduce the graph filter for the attributed graph representation learning, where the high-order neighborhood correlation is also developed for clustering.

	\subsection{Deep Attributed Graph Clustering}
	The main purpose of GNN is to learn the discriminative deep representation from the attributed graph.
	Graph Auto-Encoder~(GAE)~\cite{DBLP:journals/corr/KipfW16a}~based methods encode the adjacent matrix and attribute feature matrix into a deep representation, and then decode it with the decoder. Pan~\textit{et al.}~\cite{DBLP:conf/ijcai/PanHLJYZ18} propose an adversarial graph embedding framework for clustering, where a discriminator is employed to force the deep representation to match the prior distribution. Wang~\textit{et al.}~\cite{DBLP:conf/ijcai/WangPHLJZ19} incorporate cluster centroid learning and deep representation learning in a self-optimizing manner. In~\cite{DBLP:conf/www/Bo0SZL020}, the Graph Convolutional Network~(GCN) is introduced to extract the multi-level deep representation of the attributed graph data for clustering. Sun~\textit{et al.}~\cite{DBLP:journals/kbs/SunLDZT21} employ the dual-decoder network to reconstruct the attribute feature matrix and the adjacent matrix of the input graph simultaneously.
	The above-mentioned methods are designed for single-view attributed graph data, and the complementary information of multi-view attributed graph data may not be fully exploited without considering the diversity of the inputs.
	Inspired by~\cite{DBLP:conf/icml/XieGF16}, Cheng~\textit{et al.}~\cite{DBLP:conf/ijcai/ChengWTXG20} introduce a consistent embedding encoder to maximize the agreements between views for a consensus representation fitting the target distribution. Fan~\textit{et al.}~\cite{DBLP:conf/www/FanWSLLW20} define a most informative view with the modularity and decode the deep representation calculated from it to reconstruct multiple adjacent matrices. However, all GNN-based deep multi-view attributed graph clustering methods reconstruct the affinity matrix of each view from the consensus representation without explicitly considering the impact of the view-specific information, ultimately leading to sub-optimal clustering performance.

	\subsection{Information Maximization}
	Information Maximization~(or contrastive learning) has achieved great success in the field of computer vision and inspired many studies on unsupervised learning.
	Velickovic~\textit{et al.}~\cite{DBLP:conf/iclr/VelickovicFHLBH19} maximize the mutual information between the local patches and the global graph to obtain a compact representation. In~\cite{DBLP:conf/aaai/0003LNW0S21}, the information maximization is developed to maximize the agreement between the deep representations obtained by conducting GNN on the input graph and the augmented one.
	Pan~\textit{et al.}~\cite{DBLP:conf/nips/PanK21} employ a contrastive learning term to maximize the mutual information between each attribute node and its $k$-nearest neighbors.
	In~\cite{DBLP:conf/nips/WuRLL20}, the graph information bottleneck is proposed, which aims to balance the expressiveness and robustness of the deep representation.
	Yu~\textit{et al.}~\cite{DBLP:conf/iclr/YuXRBHH21} employ the graph information bottleneck theory to address the sub-graph recognition problem.
	Suresh~\textit{et al.}~\cite{DBLP:conf/nips/SureshLHN21} propose a graph information bottleneck-based edge augmentation module to adaptively drop the redundancy edges on the input graph. To further reduce the redundancy information between the input graph and adaptive augmented one, Gong~\textit{et al.} develop a variant by imposing the orthogonal constraint on the deep representations~\cite{DBLP:conf/ijcai/GongZTL22}.
	
	\section {Basic Notations and Problem Definition}
	\label{sec:basic notation}

	\begin{table}[!t]
		\vskip -0.15in
		\caption{Basic notations and descriptions.}
		\label{tab:notation}
		\centering
		\begin{tabular}{|c| l| }
			\hline
			Notation & Meaning\\
			\hline
			$\alpha$ &The hyper-parameter\\
			$d$  &The dimension of each attribute node\\
			$e$ &The number of layers in the GE module\\
			$r$& The number of layers in the SIR module\\
			$o$& The freedom degree of the Student's $t$-distribution\\
			$c$ &The number of clusters\\
			$\theta_l$ &The parameters of the $l$-th layer in the SIR module\\
			$\mathcal{G}$ & The attributed graph\\
			$\bm{\mathcal{G}}$ & The multi-view attributed graph\\
			$\mathcal{E}$ & The edge set\\
			$\mathcal{V}$& The node set\\
			$N$&The number of nodes\\
			$M$ &The number of edges\\
			$V$ & The number of views\\
			$\tilde{A^v}$ & The adjacent matrix\\
			$D$&The degree matrix of the adjacent matrix\\
			$I$& The identity matrix\\
			$A^v$&The normalized adjacent matrix\\
			$X$ & The attribute feature matrix\\
			
			$Z_l^v$ &The hidden representation of the $l$-th encoder layer\\
			$W_l$& The parameters of the $l$-th layer in the GE module\\
			
			$H^v$& The low-level representation\\
			$S$& The latent high-level clustering-oriented representation\\
			$R_l^v$& The reconstructed hidden representation\\
			
			$\hat{A^v_s}$&The reconstructed specific information matrix\\
			$\hat{A^v}$ &The reconstructed adjacent matrix\\
			$Q$&The soft label distribution\\
			$P$& The target distribution\\
			$\mathcal{K}$& The clustering results\\
			\hline
		\end{tabular}
	\end{table}
	
	Given an attributed graph $\mathcal{G}=\left\{\mathcal{E},\mathcal{V}\right\}$, where $\mathcal{E}$ is the edge set and $\mathcal{V}=\left\{v_1,v_2,\dots,v_N\right\}$ is the node set, the numbers of edges and nodes are denoted as $M=|\mathcal{E}|$ and $N=|\mathcal{V}|$, respectively. $X\in\mathbb{R}^{N\times d}$ is the attribute feature matrix and $d$ is the dimension of each attribute node. $\tilde{A}\in \mathbb{R}^{N\times N}$ is the adjacent matrix of $\mathcal{G}$ where $\tilde{a}_{ij}=1$ denotes that node $v_i$ and node $v_j$ are connected by an edge from $\mathcal{E}$ and vice versa. The degree matrix of $\tilde{A}$ is $D=diag\left(d_1,d_2,\dots,d_N\right)\in \mathbb{R}^{N\times N}$ and $d_i=\sum_i \tilde{a}_{ij}$. Then, $\tilde{A}$ can be further normalized as $A$ by calculating $D^{-\frac{1}{2}}\left(\tilde{A}+I\right)D^{-\frac{1}{2}}$, where $I\in \mathbb{R}^{N\times N}$ is the identity matrix. For clarity, the basic notations used throughout this paper are summarized in Table~\ref{tab:notation}.

	\textbf{Definition 1}~(Multi-view Attributed Graph):
	The multi-view attributed graph is $\bm{\mathcal{G}}=\left\{\mathcal{G}^1,\mathcal{G}^2,\dots,\mathcal{G}^V\right\}$ consisting of an attribute feature matrix $X$ and multiple adjacent matrices $\left\{\tilde{A}^{v}\right\}_{v=1}^{V}$, where $V$ is the number of views.
	
	\textbf{Definition 2}~(Multi-view Attributed Graph Clustering): Given a multi-view attributed graph, the multi-view attributed graph clustering aims to mine a consensus cluster partition fitting all views.
	
	\section{The Proposed Method}
	\begin{figure*}[!htpb]
		\centering
		\includegraphics[width=0.9\textwidth]{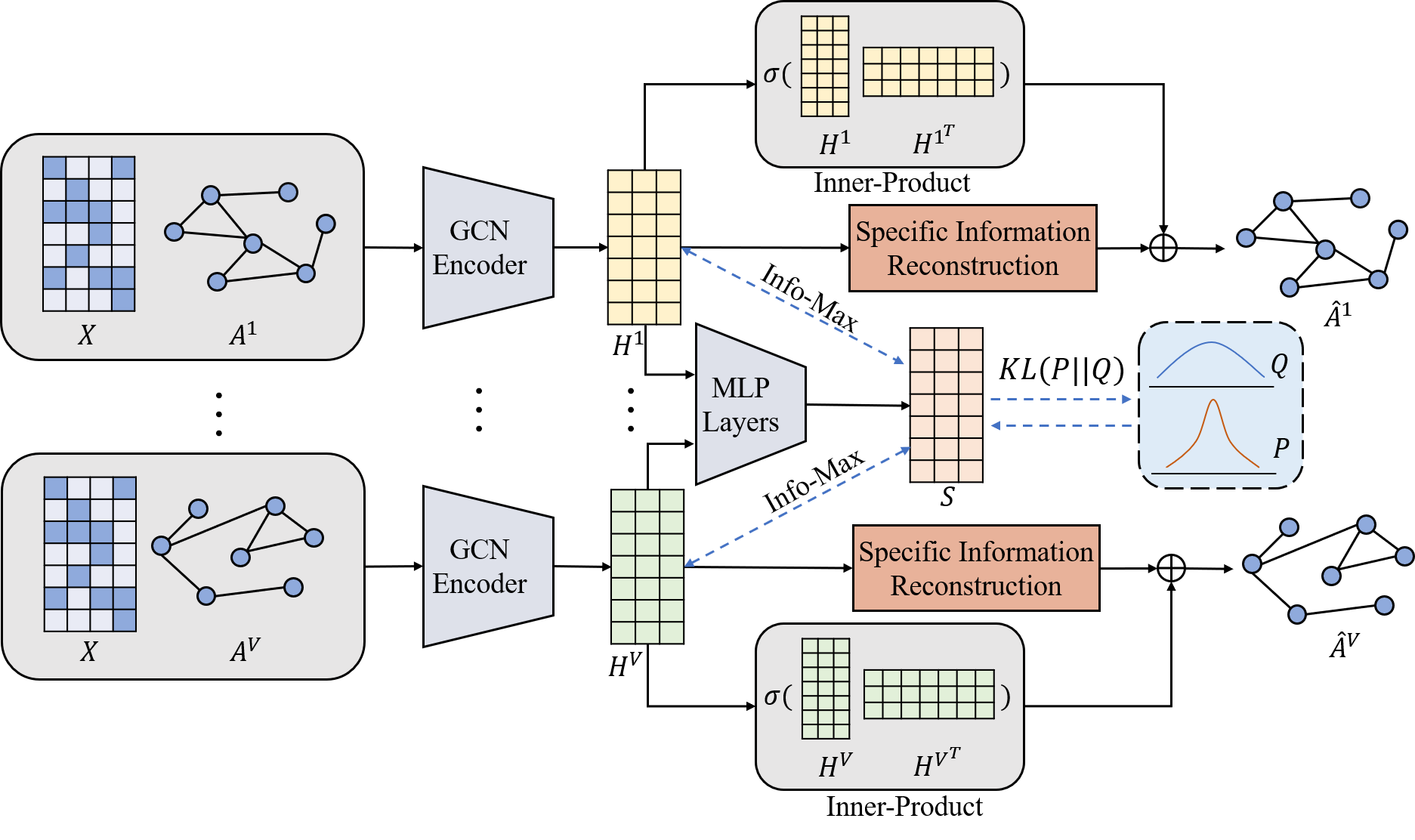}
		\caption{The framework of DIAGC. For the proposed method, we avoid the operation to obtain the clustering results from the low-level representations $H^v$, and recover the latent high-level representation $S$ by the MLP layers, where the consensus information is retained by maximizing the mutual information between $S$ and $H^v$. Meanwhile, the specific information is reconstructed by the Specific Information Reconstruction~(SIR) module disentangling the exploration of the consensus and specific information. Furthermore, the Self-supervised Clustering~(SC) module is introduced to guide $S$ to be a clustering-oriented representation.
		}
		\label{fig:framework}
	\end{figure*}
	\label{sec:The Proposed Method}
	In this section, we propose a Dual Information enhanced multi-view Attributed Graph Clustering~(DIAGC) method.

	\subsection{Motivation}
	
	Generally, most existing deep multi-view attributed graph clustering methods only focus on the consensus information between views while neglecting the specific information of each view. As illustrated in \figurename~\ref{fig:acm_ajdacent}, the adjacent matrices of the ACM dataset are distinct between different views, indicating that each view has its own specific information. For multi-view attributed graph data, except for the study of the consensus representation, how to discover the specific information of different views remains a challenging problem. In addition, these methods reconstruct the attribute graph of each view from the consensus representation, which may cause the conflict between learning the consensus representation and reconstructing the view-specific attribute graph. Specifically, the consensus objective aims to learn the common information across multiple views as much as possible while the reconstruction objective expects the common representation to preserve the view-specific information of different views. Furthermore, these methods obtain the final cluster partition directly from the low-level representations, which can not fully mine the latent high-level information, i.e., the clustering-oriented information embedded in all views.

	To this end, the basic idea of the DIAGC method is to disentangle the learning procedure of the consensus and view-specific information for acquiring a view-consistent cluster partition, and maximize the mutual information between the low-level representations and the latent high-level clustering-oriented one. For clarity, the overall framework of DIAGC is illustrated in \figurename~\ref{fig:framework}.

	\subsection{Methodology}
	The proposed DIAGC method consists of four modules, i.e., Graph Encoder~(GE) module, Mutual Information Maximization~(MIM) module, Specific Information Reconstruction~(SIR) module and Self-supervised Clustering~(SC) module.
	\subsubsection{Graph Encoder~(GE) Module}
	To capture the low-level representations of each view, the graph encoder module is introduced. Specifically, the graph encoder is utilized as a non-linearity feature extractor to mine the deep non-linearity correlations embedded in the attribute graph of each view:
	\begin{equation}
		Z^v_{l} = \sigma \left(A^v Z^v_{l-1}W_l \right),
		\label{eq:Z}
	\end{equation}
	\begin{equation}
		Z^v_{0} = X,~H^v=Z^v_{e},
	\end{equation}
	where $W_l$ denotes the parameters of the $l$-th layer, $e$ is the number of layers, $\sigma$ is the activation function, $Z_l^v$ is the hidden representation and $H^v$ is the low-level representation of the $v$-th view.
	
	\subsubsection{Mutual Information Maximization~(MIM) Module}
	Mutual information maximization~\cite{DBLP:conf/ijcai/WangPHLJZ19} encourages the same samples between different views to be similar. To explore the view-consensus high-level information embedded in each $H^v$, we employ the Mutual Information Maximization (MIM) module:
	\begin{equation}
		I\left(H^v,S\right)=\frac{1}{N}\sum_{j=1}^{N}\log \frac{exp\left(sim \left(h^v_j,s_j\right)\right)}{\sum_{j^{'}=1,j^{'}\neq j}^N exp\left(sim\left(h^v_{j},s_{j^{'}}\right)\right)},
	\end{equation}
	where $h^v_j\in \mathbb{R}^{1\times d}$ is the $j$-th row of $H^v$, and $S$ is the latent high-level representation recovered from all views. In particular, $S$ is generated by:
	\begin{equation}
		S = \text{MLP}\left([H^1;H^2;\dots;H^V]\right),
		\label{eq:S}
	\end{equation}
	where $\text{MLP}(\cdot)$ is a sub-network consisting of several fully connected layers. When maximizing the mutual information between $S$ and $H^v$, rather than simply adding $H^v$ together, $S$ can capture the intrinsic information of each view. Besides, the $\text{MLP}(\cdot)$ sub-network can also mine the nonlinear relationships between different views, which can fully utilize the nonlinear fitting ability for the latent high-level representation. Thus, we define a mutual information loss function as follows:
	\begin{equation}
		L_{I} = \sum_{v=1}^{V}I\left(H^v,S\right).
		\label{eq:MIM}
	\end{equation}
	
	\subsubsection{Specific Information Reconstruction~(SIR) Module}
	
	Reconstructing the specific information enables the MIM module to capture the more essential information that fits all views. Specifically, the specific information reconstruction module is realized by a three-layer graph neural network:
	\begin{equation}
		R_l^v = \sigma\left(A^vR^v_{l-1}\theta_l\right),
		\label{eq:R}
	\end{equation}
	\begin{equation}
		R_0^v = H^v,~\hat{A}^v_s = \sigma\left(R_{r}^vR_{r}^{v^T}\right),
	\end{equation}
	where $R_l^v$ is the reconstructed $l$-th layer hidden representation of the $v$-th view, $\theta_l$ denotes the parameters of the $l$-th layer, $r$ is the number of layers and $\hat{A}^v_s$ is the reconstructed specific information matrix of the $v$-th view.

	\subsubsection{Self-supervised Clustering~(SC) module}
	Since multi-view attributed graph clustering is an unsupervised task, the ground-truth labels are not available, which introduces difficulties in the network training. To tackle such a problem, self-supervised clustering module is employed to guide the network training. Specifically, the $KL$-divergence is utilized:
	\begin{equation}
		L_{KL}= KL\left(P||Q\right) = \sum_{i}^{N}\sum_{j}^{c}p_{ij}\log\frac{p_{ij}}{q_{ij}},
	\end{equation}
	where $Q\in \mathbb{R}^{N\times c}$ is the soft label distribution, i.e., $q_{ij}$ measures the possibility of the $i$-th instance belonging to the $j$-th cluster, and $P\in \mathbb{R}^{N\times c}$ is the target distribution. In our study, $q_{ij}$ is calculated by the Student's $t$-distribution~\cite{JMLR:v9:vandermaaten08a}:
	\begin{equation}
		q_{ij}= \frac{\left(1+\left\|s_i-\mu_j \right\|^2\right)^{-\frac{o+1}{2}}}{\sum_{j^{'}}\left(1+\left\|s_i-\mu_{j^{'}} \right\|^2\right)^{-\frac{o+1}{2}}},
		\label{Q_function}
	\end{equation}
	where $\mu=[\mu_1;\mu_2;\dots;\mu_c]\in \mathbb{R}^{c\times d}$ is the cluster centroids, 
	$o$ is the freedom degree of the Student's $t$-distribution. And the target distribution $P$ is defined as:
	\begin{equation}
		p_{ij}=\frac{q^2_{ij}/\sum_i q_{ij}}{\sum_{j^{'}}\left(q_{ij^{'}}^2 / \sum_{j^{'}}q_{ij^{'}} \right)},
	\end{equation}
	where the second power on $q_{ij}$ is utilized to make the target distribution $P$ denser. It enhances the learning process of $Q$, i.e., pushing away samples belonging to different clusters and gathering samples of the same cluster together. To initialize the cluster centroid in Eq.~(\ref{Q_function}), the $k$-means clustering method is performed on $S$ before the network training. After that, $\mu$ is updated adaptively in each training iteration. The SC module enables $S$ to be a high-level clustering-oriented representation.
		\begin{algorithm}[!t]
		\caption{Dual Information enhanced multi-view Attributed Graph Clustering~(DIAGC)}\label{alg:alg1}
		\renewcommand{\algorithmicrequire}{\textbf{Input:}}
		\renewcommand{\algorithmicensure}{\textbf{Output:}}
		\begin{algorithmic}[1]
			
			\REQUIRE Attribute feature matrix: $X\in\mathbb{R}^{N\times d}$, adjacent matrices:$\left\{\tilde{A}^v\right\}^{V}_{v=1}$, maximum iteration number $T$ and cluster number $c$.
			\ENSURE  Clustering results $\mathcal{K}$.
			\STATE \textbf{Initialize:}
			Randomly initialize the parameters of the whole network.\\
			Compute the initial cluster centroids $\mu$ based on S.
			\FOR {$t=1:T$}
			\STATE {Generate the low-level representation $H^v$ of each view by Eq.~(\ref{eq:Z}).}
			\STATE Calculate $S$ by Eq.~(\ref{eq:S}).
			\STATE Reconstruct the specific information from $H^v$ by Eq.~(\ref{eq:R}).
			\STATE Update the parameters of the DIAGC model by minimizing Eq.~(\ref{eq:total_loss}).
			\ENDFOR
			\STATE Obtain the clustering results $\mathcal{K}$ by performing the $k$-means clustering method on the latent high-level clustering-oriented representation $S$.
		\end{algorithmic}
		\label{alg1}
	\end{algorithm}

	\begin{table*}[t]
		\caption{Descriptions of the multi-view attributed graph datasets.}
		\label{tab:dataset}
		\centering
		\begin{tabular}{|ccccccc|}
			\hline
			Dataset               & \#Views            & \#Nodes               & \#Feature dimensions            & \#Edges of the adjacent matrix & Attribute content                           & \#Clusters          \\
			\hline
			\multirow{2}{*}{ACM}  & \multirow{2}{*}{2} & \multirow{2}{*}{3025} & \multirow{2}{*}{1830} & Co-Subject~(29,281)          & \multirow{2}{*}{Keywords of the paper}      & \multirow{2}{*}{3}  \\
			\cline{5-5}
			&                    &                       &                       & Co-Author~(2,210,761)        &                                             &                     \\
			\hline
			\multirow{3}{*}{DBLP} & \multirow{3}{*}{3} & \multirow{3}{*}{4057} & \multirow{3}{*}{334}  & Co-Author~(11,113)           & \multirow{3}{*}{Keywords of the author}     & \multirow{3}{*}{4}  \\
			\cline{5-5}
			&                    &                       &                       & Co-Conference~(5,000,495)    &                                             &                     \\
			\cline{5-5}
			&                    &                       &                       & Co-Term~(6,776,335)          &                                             &                     \\
			\hline
			\multirow{2}{*}{IMDB} & \multirow{2}{*}{2} & \multirow{2}{*}{4780} & \multirow{2}{*}{1232} & Co-Actor~(98,010)            & \multirow{2}{*}{Keywords of the movie plot} & \multirow{2}{*}{3}  \\
			\cline{5-5}
			&                    &                       &                       & Co-Director~(21,018)         &                                             &                     \\
			\hline
		\end{tabular}
	\end{table*}
	
	\subsection{Overall Objective Function and Algorithm Summary}
	For the network training, we jointly optimize the mutual information loss, the reconstruction loss and the self-supervised clustering loss together. The total objective function can be defined as follows:
	\begin{equation}
		L = L_{I} + L_R +\alpha L_{KL},
		\label{eq:total_loss}
	\end{equation}
	where $\alpha$ is a hyper-parameter and $L_R$ is the reconstruction loss. In our study, $L_R$ is defined as:
	\begin{equation}
		L_R = \sum_{v}^{V}\left\|A^v-\hat{A}^v \right\|_{F}^{2},
		\label{eq:L_R}
	\end{equation}
	\vskip -3mm
	 \noindent where $\hat{A}^v$ is the reconstructed adjacent matrix of the $v$-th view. $\hat{A}^v$ is calculated by:
	\begin{equation}
		\hat{A}^v = \hat{A}^v_c+\hat{A}^v_s,
	\end{equation}
	where the first term $\hat{A}^v_c$ is to capture the consensus information of the low-level representations, while the second term $\hat{A}^v_s$ aims to reconstruct the view-specific information. $\hat{A}^v_c$ is defined as:
	\begin{equation}
		\hat{A}^v_c = \sigma\left(H^vH^{v^T}\right).
	\end{equation}
	
	It is worth noting that the reason for reconstructing $\hat{A}^v_c$ from $H^v$ instead of $S$ is to ensure that $S$ is free from the interference of specific information, which may lead to sub-optimal solutions.
	With the help of Eq.~(\ref{eq:MIM}) and Eq.~(\ref{eq:L_R}), the exploration of the consensus and specific information  are disentangled elegantly, i.e., the MIM module ensures $H^v$ contains the view-common information while the SIR module reconstructs the features that are characteristic of the $v$-th view. After $T$ iterations, the final clustering results $\mathcal{K}$ are obtained by performing the $k$-means clustering method on $S$.
	
	For clarity, the overall algorithm of the proposed DIAGC method is outlined in Algorithm~\ref{alg:alg1}.

	\section{Experiments}
	\label{sec:Experiments}
	
	In this section, extensive experiments are conducted to validate the effectiveness of the proposed DIAGC method. Three widely used real-world multi-view attributed graph datasets
	and thirteen state-of-the-art baselines are adopted for the experiment.
	All experiments are conduced in Python 3.6 64-bit edition on an Intel(R) Xeon(R) Gold 6248R CPU @3.0GHz Ubuntu 20.04 workstation with 256 GB of RAM and RTX 3090 GPU.

	\subsection{Dataset Description}
	The following three datasets are used in the experiments.
	\begin{enumerate}
		\item
		\textbf{ACM}~\cite{DBLP:conf/kdd/TangZYLZS08}: It is a paper network dataset containing 3025 papers as the attribute feature matrix. Besides, two kinds of relationships, i.e., co-paper and co-author relationships, are regarded as two views of the adjacent matrices, which can be divided into three categories\footnote{ https://dl.acm.org/.}.
		\item
		\textbf{DBLP}~\cite{DBLP:conf/ijcai/PanWZZW16}: It is an author network dataset consisting of 4057 authors with three kinds of relationships, i.e., co-author, co-term and co-conference, as three different views of the adjacent matrices.
		Four categories of papers are collected: machine learning, information retrieval, database and data mining\footnote{ https://dblp.uni-trier.de/.}.
		\item
		\textbf{IMDB}~\cite{DBLP:conf/www/FanWSLLW20}: It is a movie network dataset that contains 4780 movies with two kinds of relationships, i.e., co-actor and co-director. The movies are divided into three categories: drama, action and comedy\footnote{https://www.imdb.com/.}.
	\end{enumerate}
	
	The statistical information of these datasets is summarized in Table~\ref{tab:dataset}.
		
	\subsection{Baselines}
	
	Thirteen clustering methods are compared with the proposed method. The details of these methods are provided as follows:
	\begin{enumerate}
		\item \textbf{LINE}~\cite{DBLP:conf/www/TangQWZYM15}: It is a deep single-view attributed graph clustering method, which is designed to preserve the first-order and the second-order proprieties of the attributed graph data.
		In our study, its best clustering results among all views are reported with LINE$_{best}$ and its average clustering results of all views reported with LINE$_{avg}$\footnote{https://github.com/tangjianpku/LINE.}.

		\item
		\textbf{MNE}~\cite{DBLP:conf/ijcai/ZhangQYS18}: It is a scalable multi-view network embedding model representing the adjacent relationships from different views of multi-view attributed graph data into a unified embedding space\footnote{https://github.com/HKUST-KnowComp/MNE.}.
		\item
		\textbf{PMNE}~\cite{DBLP:conf/icdm/LiuCYSC17}: The Project Multi-layer Network Embedding (PMNE) method consists of three sub-methods, i.e., network aggregation-based method~(PMNE$_n$), result aggregation-based method~(PMNE$_r$) and layer co-analysis-based method~(PMNE$_{lc}$). These methods project the multi-layer network into a delegated vector space for clustering.
		\item \textbf{GAE}~\cite{DBLP:journals/corr/KipfW16a}: The Graph Auto-Encoder~(GAE) is one of the most representative methods of single-view attributed graph clustering, which is composed of one graph encoder and one graph decoder. The best clustering results of GAE among all views are reported with GAE$_{best}$ and its average clustering results of all views are reported as GAE$_{avg}$\footnote{https://github.com/tkipf/gae.}.
		\item
		\textbf{ARGA}~\cite{DBLP:conf/ijcai/PanHLJYZ18}: The Adversarially Regularized Graph Autoencoder~(ARGA) method is a single-view attributed graph clustering method, where a discriminator is utilized to ensure the deep representation calculated by encoder matching a prior distribution\footnote{https://github.com/GRAND-Lab/ARGA}.
		\item
		\textbf{DEAGC}~\cite{DBLP:conf/ijcai/WangPHLJZ19}: It is a goal-directed deep single-view attributed graph clustering method utilizing a self-optimizing module to learn a clustering-oriented deep representation.
		\item
		\textbf{SDCN}~\cite{DBLP:conf/www/Bo0SZL020}: The Structural Deep Clustering Network~(SDCN) method integrates the deep representations extracted by two deep neural networks to recover the distinct representations embedded in the single-view data\footnote{https://github.com/bdy9527/SDCN}.
		
		\item
		\textbf{RMSC}~\cite{DBLP:conf/aaai/XiaPDY14}: It is a Markov-chain-based multi-view clustering method which takes each individual adjacent matrix as the transition probability matrix, and then obtains the shared low-rank representation from them\footnote{https://github.com/frash1989/ELM-MVClustering/tree/master/RMSC-ELM.}.
		\item
		\textbf{PwMC}~\cite{DBLP:conf/ijcai/NieLL17}: It is a multi-view clustering method that adaptively learns the weights of each view for a consensus representation with low-rank constraint\footnote{https://github.com/kylejingli/SwMC-IJCAI17/blob/master/SwMC}.
		\item
		\textbf{SwMC}~\cite{DBLP:conf/ijcai/NieLL17}: It is a self-weighted multi-view clustering method introducing a self-conducted weight learning scheme to remove the explicitly defined weight factor.
		\item
		\textbf{O2MA}~\cite{DBLP:conf/www/FanWSLLW20}: It is a deep multi-view attributed graph clustering method, which learns the deep representation from a most informative graph view and reconstructs all views from the deep representation\footnote{https://github.com/googlebaba/WWW2020-O2MAC.}.
		\item
		\textbf{O2MAC}~\cite{DBLP:conf/www/FanWSLLW20}: It is a variant of O2MA containing a self-supervised clustering module with the clustering loss to generate the cluster-orient deep representation.
		\item
		\textbf{MvAGC}~\cite{DBLP:conf/ijcai/LinK21}: It is a multi-view attributed graph clustering method which adopts the graph filter learning term to capture the high-order neighborhood information\footnote{https://github.com/sckangz/MvAGC.}.
		
	\end{enumerate}

	For fair comparison, similar to~\cite{DBLP:conf/ijcai/LinK21}, part of the clustering results of the comparison methods are copied from~\cite{DBLP:conf/www/FanWSLLW20}.

	\subsection{Evaluation Metrics}
	Four widely used evaluation metrics, namely, ACCuracy~(ACC), F1-score~(F1), Normalized Mutual Information~(NMI) and Adjusted Rand Index~(ARI)~\cite{DBLP:conf/www/FanWSLLW20} are employed to evaluate the clustering performance.
	Let $Y=\left\{Y_1,Y_2,\dots,Y_c\right\}\in \mathbb{R}^{n \times c}$ denote the ground-truth cluster partition, where $Y_i$ consists of the nodes belonging to the $i$-th cluster. Assuming that the cluster partition obtained by the clustering method is $Y^*=\left\{Y^*_1,Y^*_2,\dots,Y^*_c\right\}\in \mathbb{R}^{n \times c}$, ACC is calculated by:
	\begin{equation}
		\text{ACC} = \frac{\sum_{i=1}^c\left|Y_i \cap Y_i^* \right|}{N},
	\end{equation}
	where $\left|\cdot\right|$ is the number of the elements of the set.
	
	F1 is obtained by:
	\begin{equation}
		\text{F1} = 2\frac{Pr \cdot Re}{Pr+Re},
	\end{equation}
	where $Pr$ and $Re$ are precision and recall. They can be calculated by the following formulas:
	\begin{equation}
		Pr = \frac{TP}{TP+FP}, ~Re = \frac{TP}{TP+FN},
	\end{equation}
	where $TP$ denotes the number of true positives, $FP$ denotes the number of false positives and $FN$ denotes the number of false negatives.
	
	NMI is calculated by:
	\begin{equation}
		\text{NMI}=\frac{ -2\sum_{i,j=1}^{c} \left|Y_i \cap Y_j^*\right| \log\left(\frac{N \left|Y_i \cap Y_j^* \right|}{\left|Y_i\right|\left|Y_j^*\right|} \right) }{\sum_{i=1}^{c}\left|Y_i \right|\log\frac{\left|Y_i\right|}{N}+\sum_{j=1}^c \left|C_j^*\right|\log\frac{\left|Y^*_j\right|}{N} }.
	\end{equation}
	
	The formulation of ARI can be written as:
	\begin{equation}
		\text{ARI} = \frac{RI-E(RI)}{max(RI)-E(RI)},
	\end{equation}
	where $E(\cdot)$ is the expected value, $RI$ denotes the rand index:
	\begin{equation}
		RI = \frac{TP+TN}{TP+FP+FN+TN},
	\end{equation}
	where $TN$ is the number of true negatives. Larger ACC, F1, NMI and ARI indicate better clustering performance.

	\begin{table}[!t]
		\setlength\tabcolsep{12pt}
		\caption{Comparison results on the ACM dataset. The best results are highlighted in bold while the second-best ones in underlined. }
		\centering
		\begin{tabular}{|c|cccc|}
			\hline
			Dateset  & \multicolumn{4}{c|}{ACM}               \\
			\hline
			Method   & ACC    & F1     & NMI    & ARI         \\
			\hline
			LINE$_{best}$      & 0.6479 & 0.6595 & 0.3941 & 0.3433 \\
			LINE$_{avg}$ & 0.6479 & 0.6595 & 0.3941 & 0.3432     \\
			MNE      & 0.6370 & 0.6479 & 0.2999 & 0.2486    \\
			PMNE$_n$ & 0.6936 & 0.6955 & 0.4648 & 0.4302    \\
			PMNE$_r$ & 0.6492 & 0.6618 & 0.4063 & 0.3453    \\
			PMNE$_{lc}$ & 0.6998 & 0.7003 & 0.4755 & 0.4431   \\
			\hline
			GAE$_{best}$      & 0.8216 & 0.8225 & 0.4914 & 0.5444   \\
			GAE$_{avg}$  & 0.6990 & 0.7025 & 0.4771 & 0.4378   \\
			ARGA     & 0.8433 & 0.8451 & 0.5454 & 0.6064    \\
			DAEGC    & 0.8694 & 0.8707 & 0.5618 & 0.5935       \\
			SDCN     & \underline{0.9045} & 0.9042 & 0.6831 & 0.7391   \\
			
			\hline
			RMSC     & 0.6315 & 0.5746 & 0.3973 & 0.3312   \\
			PwMC     & 0.4162 & 0.3783 & 0.0332 & 0.0395    \\
			SwMC     & 0.3831 & 0.4709 & 0.0838 & 0.0187   \\
			\hline
			O2MA     & 0.8880 & 0.8894 & 0.6515 & 0.6987   \\
			O2MAC    & 0.9042 & \underline{0.9053} & \underline{0.6923} & \underline{0.7394}   \\
			MvAGC    & 0.8975 & 0.8986 & 0.6735 & 0.7212   \\
			DIAGC     & \textbf{0.9170}  & \textbf{0.9177} & \textbf{0.7161} & \textbf{0.7697}   \\
			\hline
		\end{tabular}
		\label{tab:comparison_1}
	\end{table}

	\begin{table}[!t]
		\setlength\tabcolsep{12pt}
		\caption{Comparison results on the DBLP dataset. The best results are highlighted in bold while the second-best ones in underlined. }
		\centering
		\begin{tabular}{|c|cccc|}
			\hline
			Dateset  & \multicolumn{4}{c|}{DBLP}         \\
			\hline
			Method   & ACC    & F1     & NMI    & ARI           \\
			\hline
			LINE$_{best}$       & 0.8689 & 0.8546 & 0.6676 & 0.6988     \\
			LINE$_{avg}$  & 0.8750 & 0.8660 & 0.6681 & 0.7056   \\
			MNE       & -      & -      & -      & -        \\
			PMNE$_n$  & 0.7925 & 0.7966 & 0.5914 & 0.5265  \\
			PMNE$_r$  & 0.3835 & 0.3688 & 0.0872 & 0.0689 \\
			PMNE$_{lc}$  & -      & -      & -      & -        \\
			\hline
			GAE$_{best}$       & 0.8859 & 0.8743 & 0.6925 & 0.7410   \\
			GAE$_{avg}$   & 0.5558 & 0.5418 & 0.3072 & 0.2577  \\
			ARGA     & 0.5816 & 0.5938 & 0.2951 & 0.2392      \\
			DAEGC    & 0.6205 & 0.6175 & 0.3249 & 0.2103      \\
			SDCN      & 0.6805 & 0.6771 & 0.3950 & 0.3915   \\
			
			\hline
			RMSC      & 0.8994 & 0.8248 & 0.7111 & 0.7647   \\
			PwMC      & 0.3253 & 0.2808 & 0.0190 & 0.0159   \\
			SwMC      & 0.6538 & 0.5602 & 0.3760 & 0.3800  \\
			\hline
			O2MA      & 0.9040 & 0.8976 & 0.7257 & 0.7705  \\
			O2MAC     & 0.9074 & 0.9013 & 0.7287 & 0.7780  \\
			MvAGC     & \underline{0.9277} & \underline{0.9225} & \underline{0.7727} & \underline{0.8276} \\
			DIAGC    & \textbf{0.9320} & \textbf{0.9273} & \textbf{0.7811} & \textbf{0.8357}   \\
			\hline
		\end{tabular}
		\label{tab:comparison_2}
	\end{table}

	\begin{table}[!t]
		\setlength\tabcolsep{12pt}
		\caption{Comparison results on the IMDB dataset. The best results are highlighted in bold while the second-best ones in underlined. }
		\centering
		\begin{tabular}{|c|cccc|}
			\hline
			Dateset        & \multicolumn{4}{c|}{IMDB}          \\
			\hline
			Method       & ACC    & F1     & NMI    & ARI     \\
			\hline
			LINE$_{best}$      & 0.4268 & 0.2870  & 0.0031 & -       \\
			LINE$_{avg}$  & 0.4719 & 0.2985 & 0.0063 & -       \\
			MNE       &0.3958 & 0.3316 & 0.0017 & 0.0008  \\
			PMNE$_n$  & 0.4958 & 0.3906 & 0.0359 & 0.0366  \\
			PMNE$_r$  & 0.4697 & 0.3183 & 0.0014 & 0.0115  \\
			PMNE$_{lc}$  & 0.4719 & 0.3882 & 0.0285 & 0.0284  \\
			\hline
			GAE$_{best}$      & 0.4298 & 0.4062 & 0.0402 & 0.0473  \\
			GAE$_{avg}$   & 0.4442 & 0.4172 & 0.0413 & 0.0491  \\
			ARGA     & 0.4119 & 0.3685 & 0.0063 & -       \\
			DAEGC     & -      & -      & -      & -       \\
			SDCN      & 0.4697 & 0.3183 & 0.0285 & 0.0284  \\
			\hline
			RMSC      & 0.2702 & 0.3775 & 0.0054 & 0.0018  \\
			PwMC      & 0.2453 & 0.3164 & 0.0023 & 0.0017  \\
			SwMC      & 0.2671 & 0.3714 & 0.0056 & 0.0004  \\
			\hline
			O2MA      & 0.4697 & \underline{0.4229} & \underline{0.0524} & 0.0753  \\
			O2MAC     & 0.4502 & 0.4159 & 0.0421 & 0.0564  \\
			MvAGC     & \underline{0.5633} & 0.3783 & 0.0371 & \underline{0.0940}  \\
			DIAGC      & \textbf{0.5839} & \textbf{0.4301} & \textbf{0.0658} & \textbf{0.1316}  \\
			\hline
		\end{tabular}
		\label{tab:comparison_3}
	\end{table}
	
	\subsection{Comparison Experiments}
	
	In the comparison experiments, we run each method 10 times and report their average values in terms of ACC, F1, NMI and ARI in Table~\ref{tab:comparison_1}, Table~\ref{tab:comparison_2} and Table~\ref{tab:comparison_3}. From these tables, we have the following observations:
	\begin{enumerate}
		\item The proposed DIAGC method achieves the best clustering performance compared with the baselines. These results verify that DIAGC could leverage the complementary information between different views to generate a more comprehensive representation for clustering.
		\item Compared with the earlier multi-view clustering methods, i.e., RMSC, PwMC and SwMC, 
		most of deep learning-based methods achieve a large margin in terms of the evaluation metrics on the three multi-view attributed graph datasets. The results indicate that exploring the non-linearity relationships embedded in data is crucial for learning more discriminative representation.
		\item The GCN-based single-view clustering methods, i.e., GAE, AGRA, DAEGC and SDCN, perform much better than other single-view methods. The reason may be that GCN could incorporate both structural information embedded in the adjacent matrix and attribute information to mine the essential representation of attributed graph data.
		
		\item O2MA and O2MAC are both GCN-based multi-view attributed graph clustering methods. They utilize the consensus representation generated by GCNs as the deep representation for clustering and reconstruct adjacent matrices of all views from it. However, these two methods ignore the specific information of each view, which results in sub-optimal clustering performance.
		From Table~\ref{tab:comparison_1}, it can be seen that the proposed DIAGC method performs much better than these two methods on all the benchmarks. Taking the ACM dataset as an example, it achieves 6.46\% and 2.38\% improvements in terms of NMI than O2MA and O2MAC, respectively. The reason is that the proposed DIAGC method could disentangle the consensus and the specific information learning procedures of each view. Such a process enables the network to generate a more clustering-friendly representation by eliminating the influence caused by the inherent specific information embedded in each view. With the help of specific information reconstruction module, the proposed DIAGC method could outperform all the deep baselines on the three datasets.
		\item On the IMDB dataset, all the baselines do not perform very well, which is caused by its sparsity. Even so, the proposed method can still achieve the best clustering performance on the IMDB dataset, which further validates its effectiveness.
	\end{enumerate}
	
	In conclusion, the proposed DIAGC method can achieve more accurate clustering results than the existing state-of-the-art methods.
	\begin{table}[!t]
		\caption{The ablation study about MIM, SIR and SC module on all the benchmarks. The best results are highlighted in bold.}
		\centering
		\begin{tabular}{|c|ccc|cccc|}
			\hline
			Dateset               & MIM & SIR & SC & ACC    & F1     & NMI    & ARI     \\
			\hline
			\multirow{4}{*}{ACM}
			&    & \checkmark  & \checkmark  & 0.3901 & 0.3114 & 0.0609 & 0.007   \\
			\cline{2-8}
			& \checkmark   &    & \checkmark & 0.7709 & 0.7719 & 0.5117 & 0.4975  \\
			\cline{2-8}
			& \checkmark   & \checkmark   &   & 0.8955 & 0.8963 & 0.6603 & 0.7167  \\
			\cline{2-8}
			& \checkmark   & \checkmark   & \checkmark  & \textbf{0.9170} & \textbf{0.9177} & \textbf{0.7161} & \textbf{0.7697}  \\
			\hline
			\multirow{4}{*}{DBLP} &   & \checkmark   & \checkmark & 0.2916 & 0.1909 & 0.0268 & 0.0010   \\
			\cline{2-8}
			& \checkmark   &   & \checkmark  & 0.9295 & 0.9254 & 0.7706 & 0.8290   \\
			\cline{2-8}
			& \checkmark   & \checkmark   &   & 0.9300   & 0.9255 & 0.7708 & 0.8309  \\
			\cline{2-8}
			& \checkmark   & \checkmark   & \checkmark  & \textbf{0.9320}  & \textbf{0.9273} & \textbf{0.7811} & \textbf{0.8357}  \\
			\hline
			\multirow{4}{*}{IMDB}
			&    & \checkmark   & \checkmark  & 0.5423 & 0.2614 & 0.0022 & 0.0089  \\
			\cline{2-8}
			& \checkmark   &   & \checkmark  & 0.5073 & 0.4238 & 0.0509 & 0.1049  \\
			\cline{2-8}
			& \checkmark   & \checkmark   &   & 0.4473 & 0.4221 & 0.0561 & 0.0650   \\
			\cline{2-8}
			& \checkmark   & \checkmark   & \checkmark  & \textbf{0.5839} & \textbf{0.4301} & \textbf{0.0658} & \textbf{0.1316}  \\
			\hline
		\end{tabular}
		\label{tab:ablation}
	\end{table}
	\begin{figure}[!t]
		\renewcommand{\subfigcapskip}{0pt}
		\renewcommand{\subfigbottomskip}{0pt}
		\centerline
		{
			{\subfigure
				{\includegraphics[width=0.25\textwidth]{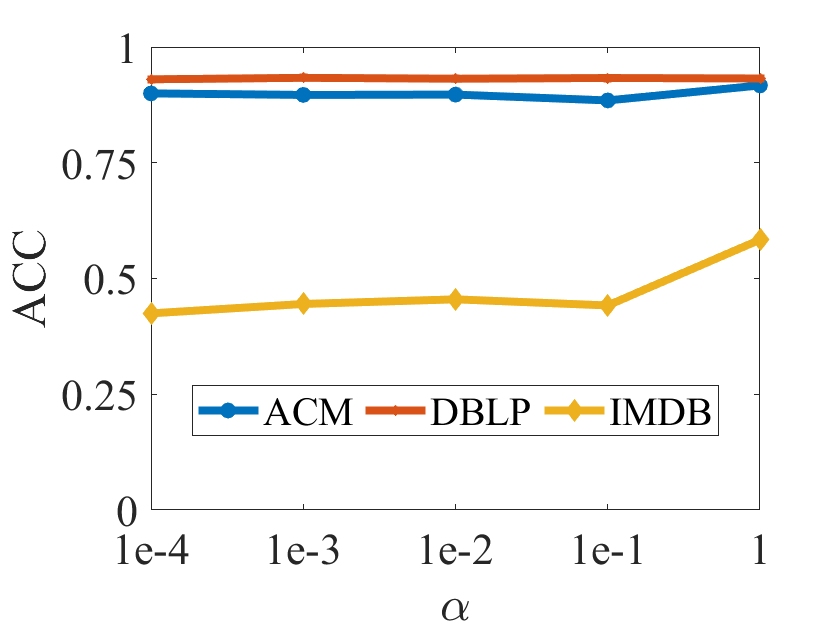}
					\label{fig:acc_param}
			}}

			{\subfigure
				{\includegraphics[width=0.25\textwidth]{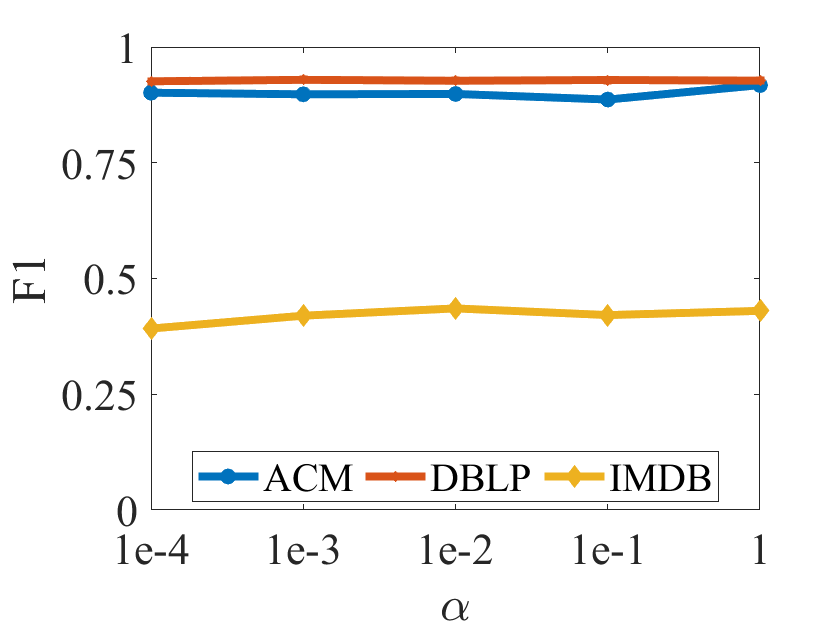}
					\label{fig:f1_param}
			}}
		}
		\centerline
		{
			{\subfigure
				{\includegraphics[width=0.25\textwidth]{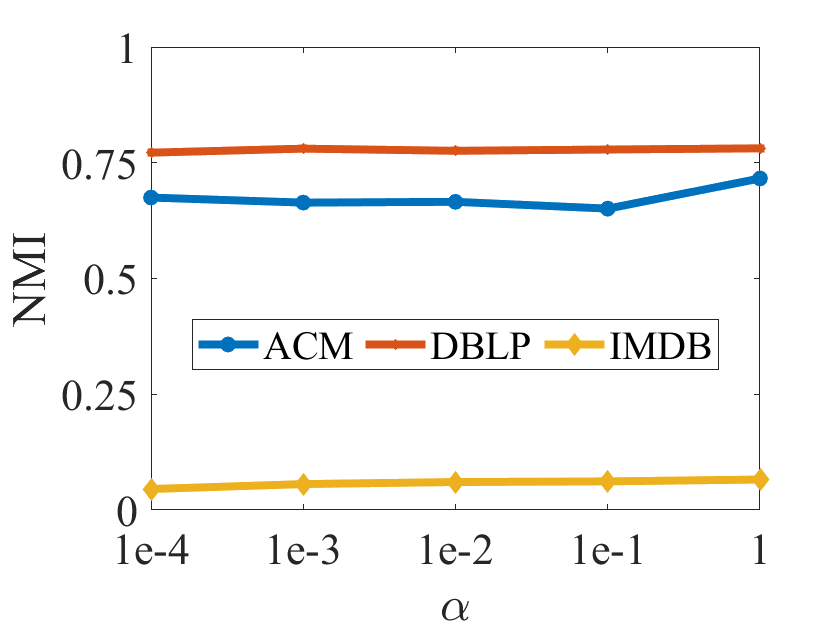}
					\label{fig:nmi_param}
			}}

			{\subfigure
				{\includegraphics[width=0.25\textwidth]{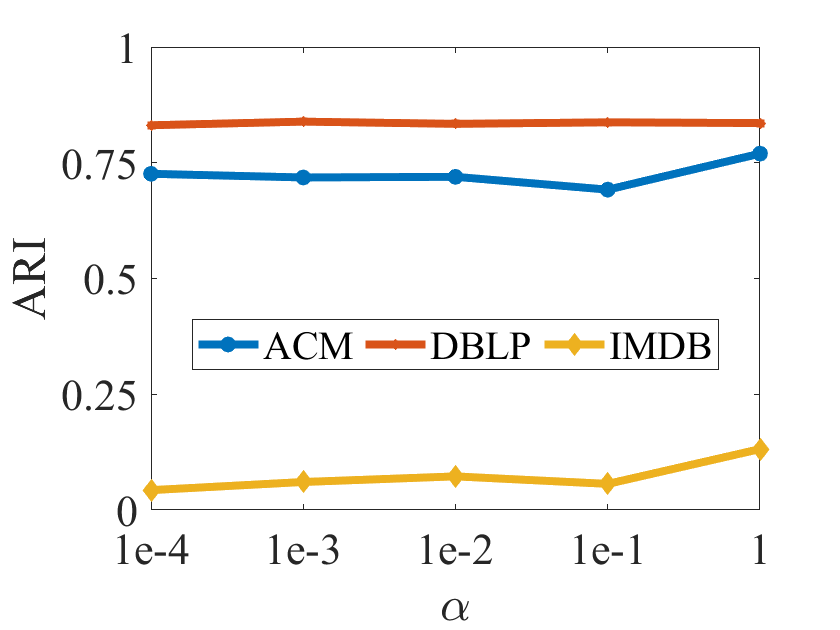}
					\label{fig:ari_param}
			}}
			
		}
		\caption{The parameter analysis of $\alpha$ on all the benchmarks.}
		\label{fig:param}
	\end{figure}
	\begin{figure*}[!t]
		\renewcommand{\subfigcapskip}{0pt}
		\renewcommand{\subfigbottomskip}{0pt}
		\centerline
		{
			{\subfigure[DEAGC]
				{\includegraphics[width=0.33\textwidth]{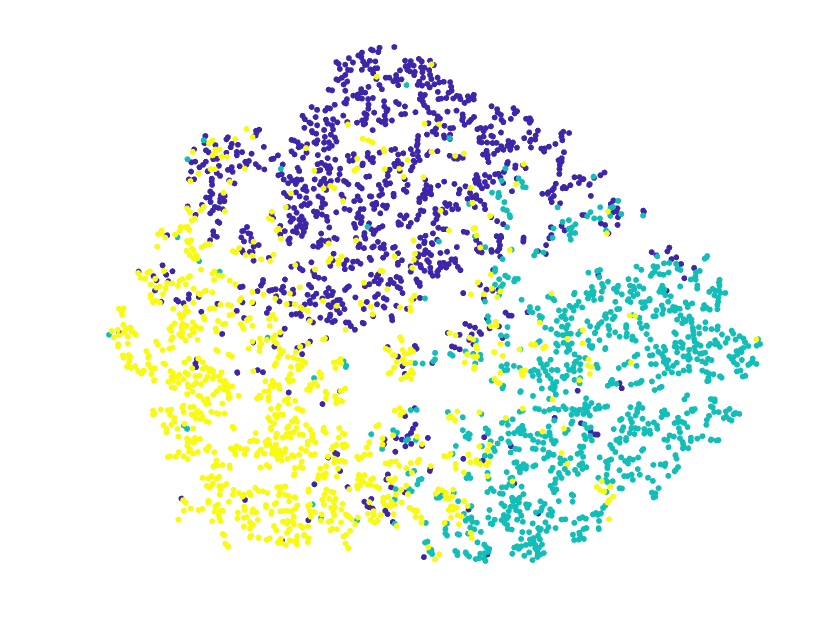}
					\label{fig:DEAGC_vis}
			}}

			{\subfigure[O2MAC]
				{\includegraphics[width=0.33\textwidth]{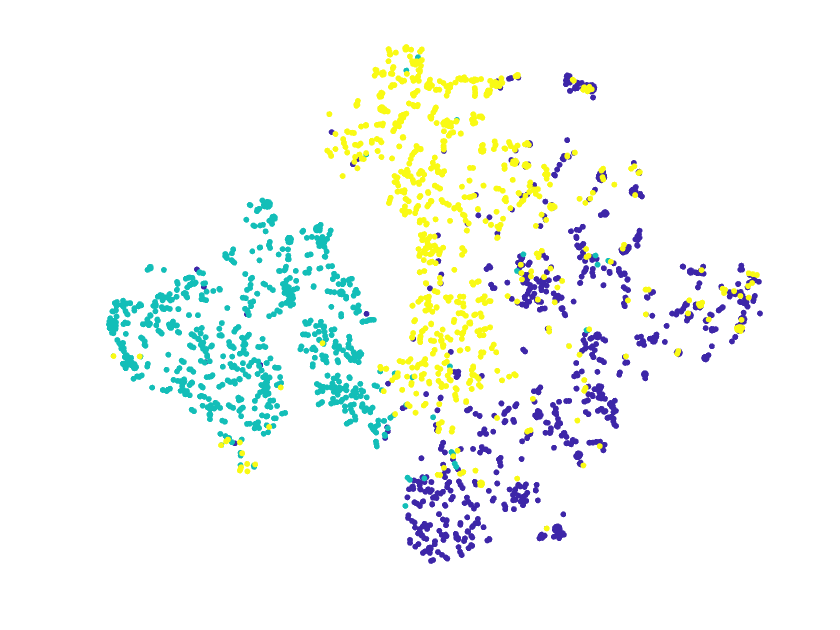}
					\label{fig:O2MAC_vis}
			}}
			
			{\subfigure[DIAGC]
				{\includegraphics[width=0.33\textwidth]{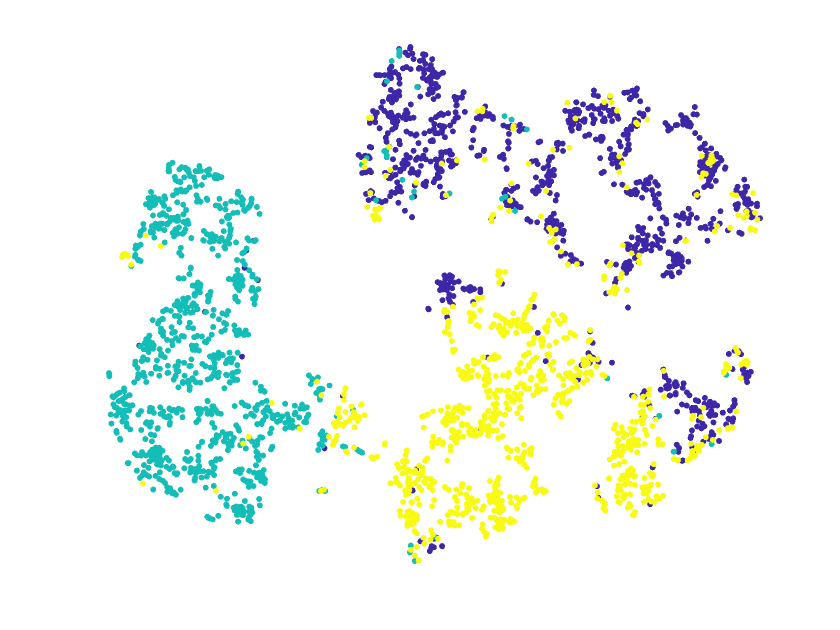}
					\label{fig:Ours_vis}
			}}
		}	
		\caption{The t-SNE visualization on the ACM dataset. }
		\label{fig:vis}
	\end{figure*}
	\begin{figure*}[!t]
		\vskip -0.1in
		\renewcommand{\subfigcapskip}{0pt}
		\renewcommand{\subfigbottomskip}{0pt}
		\centerline
		{
			{\subfigure[ACM]
				{\includegraphics[width=0.33\textwidth]{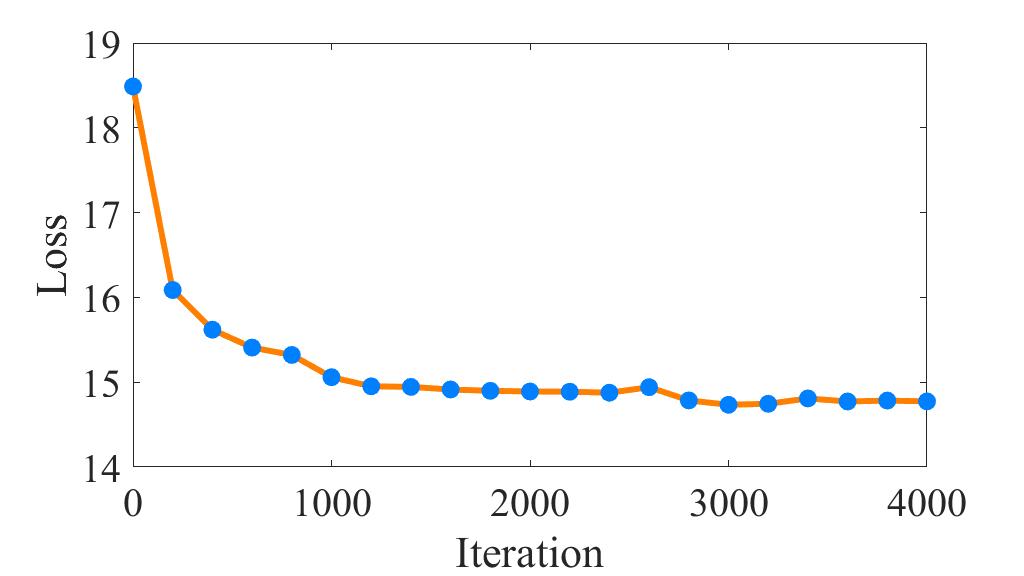}
					\label{fig:acm_converge}
				}	
			}
			{\subfigure[DBLP]
				{\includegraphics[width=0.33\textwidth]{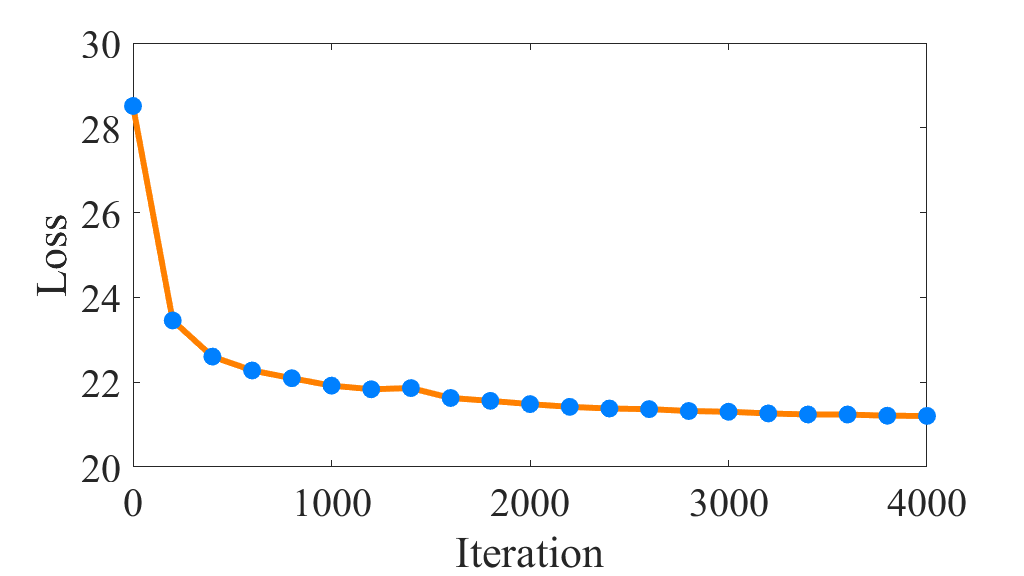}
					\label{fig:dblp_converge}
				}
			}
			{\subfigure[IMDB]
				{\includegraphics[width=0.33\textwidth]{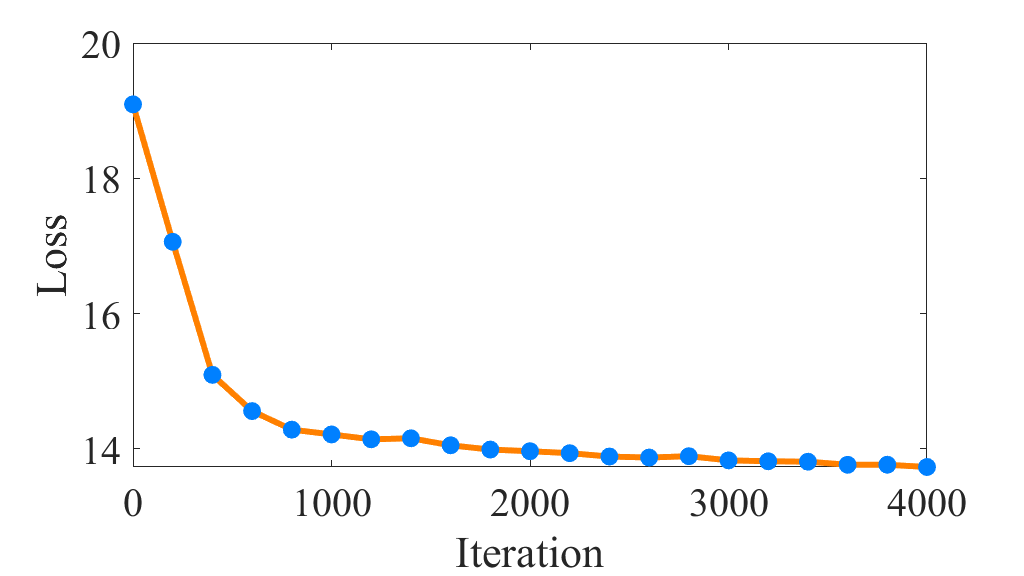}
					\label{fig:imdb_converge}
				}
			}	
		}
		\caption{The convergence curves of all the benchmarks.}
		\label{fig:converge}
		\vskip -0.15in
	\end{figure*}
	
	\subsection{Ablation Study}
	To further validate the effectiveness of MIM, SIR and SC modules of the proposed DIAGC method, in this subsection, the ablation study is conducted and the experiment results are reported in Table~\ref{tab:ablation}. It is noteworthy that without the MIM module, the SC module is performed on the representation generated by the linear sum of all the deep representations, i.e., the SC module is performed on $S^{'}=\frac{1}{V}\sum_{v=1}^V H^v$, and the final clustering results are obtained by performing the $k$-means on $S^{'}$. Without SIR module, the $L_R$ in Eq.~(\ref{eq:total_loss}) is reformulated as:
	\begin{equation}
		L_{R^{'}}=\sum_{v}^{V}\left\|A^v-\sigma\left(H^vH^{v^T}\right) \right\|_F^2.
	\end{equation}
	
	From Table~\ref{tab:ablation}, it can be seen that without the MIM module, the clustering results of DIAGC drop dramatically on all the benchmarks. The reason may be that $S^{'}$ is generated by linearly summing the deep representations of all views with the same weight, and such a naive learning strategy breaks the non-linear structure constructed by GNN, where there are only the low-level representations.
	In addition, the clustering results for all benchmarks are degraded in the absence of the SIR module, because the low-level representations $H^v$ computed by GCN without the SIR module are disturbed by the inherent specific information embedded in each view.
	Moreover, the effectiveness of the SC module is also validated and it improves the clustering results in terms of NMI by 5.58\%, 1.03\% and 0.97\% on the ACM, DBLP and IMDB datasets, respectively. These results demonstrate that all the modules in the proposed DIAGC method are essential to improve the clustering performance.
	

	\subsection{Parameter Analysis}
	In this subsection, we would conduct the parameter analysis on all the benchmarks with the hyper-parameter $\alpha$, which is tuned in the range of $\{0.0001,0.001,0.01,0.1,1\}$. \figurename~\ref{fig:param} plots the clustering results in terms of ACC, F1, NMI and ARI on all the benchmarks, respectively. According to \figurename~\ref{fig:param}, we can observe that the clustering performance of DIAGC is relatively stable over all the benchmarks with different values of $\alpha$, which indicates the robustness of DIAGC to obtain the promising clustering performance.

	\subsection{Visualization}
	The visualization experiments of the proposed method compared with some baselines will be presented in this subsection to validate the superiority of the proposed DIAGC method. Specifically, we project the deep representations learned from the different models into a two-dimensional space. Then, the t-SNE visualization method is performed with all the representations on the ACM dataset. As shown in \figurename~\ref{fig:vis}, the DEAGC method does not perform well on the ACM dataset as it could not divide the data appropriately. The O2MAC method could separate the green data points from the rest of the data, but the other clusters are still blurry. One reason may be that the O2MAC method can only encode one view to calculate the deep representation, without considering the complementary information from different views. From \figurename~\ref{fig:Ours_vis}, it can be seen that the cluster structures obtained by DIAGC are generally clear, which validates the effectiveness of the proposed method.

	\subsection{Convergence Analysis}
	
	In this subsection, the convergence property of the proposed DIAGC method would be verified. Specifically, we plot the convergence curves about Eq.~(\ref{eq:total_loss}) versus iterations in \figurename~{\ref{fig:converge}}.
	As shown in \figurename~\ref{fig:converge}, the convergence curves are generally monotonically decreasing except for the small amplitude fluctuation caused by $L_{KL}$. Since the clustering centroids of $Q$ are recalculated at each iteration, some bias would be inevitably introduced, resulting in that the $L_{KL}$ between the soft label distribution and the target distribution is not always decreasing. Even though, the overall convergence curves of the proposed DIAGC method are decreasing and remain steady on all the benchmarks, which indicates the desirable convergence property of the proposed method.

	\section{Conclusions}
	\label{sec:Conclusion}
	In this paper, a novel Dual Information enhanced multi-view Attributed Graph Clustering~(DIAGC) method is proposed. To the best of our knowledge, this is the first attempt to disentangle the consensus and specific information learning. Specifically, the proposed DIAGC method introduces the specific information reconstruction module to disentangle the exploration of the consensus and specific information, which enables the GCN encoder to calculate the more essential low-level representations.
	Further, the latent high-level representation is recovered from the low-level ones, and the consensus information of all views is retained by the mutual information maximization module. Finally, the self-supervised clustering module is employed to enforce the high-level representation to be the clustering-oriented one.
	The experimental results on several real-world benchmarks demonstrate the effectiveness of the proposed method.

	
	\bibliographystyle{IEEEtran}

\end{document}